\documentclass[11pt]{article}
\usepackage[margin=1in]{geometry}
\usepackage{setspace}

\usepackage{amsmath,amssymb}
\usepackage{graphicx}
\usepackage{hyperref}
\usepackage{authblk}
\usepackage{xcolor}
\usepackage[normalem]{ulem} % strike through text with \sout{...}
\usepackage{url}

\title{Why AI Alignment Failure Is Structural:\\
Learned Human Interaction Structures and AGI as an Endogenous Evolutionary Shock}

\author[1]{Didier Sornette}
\author[1]{Sandro Claudio Lera}
\author[1]{Ke Wu}
\affil[1]{Institute of Risk Analysis, Prediction and Management (Risks-X), Southern University of Science and Technology (SUSTech), Shenzhen, 518055, China}
\date{}
\begin{document}
\maketitle

\begin{quote}
{\small
Richard Feyman (26th Sept. 1985): {\it I think that we are getting close to intelligent machines but they're showing the necessary weaknesses of intelligence.}\\
 Interview: \url{https://www.youtube.com/watch?v=ipRvjS7q1DI} \\(the quote is said around 17:33 minutes mark.)
 }
 \end{quote}
 
%%%%%%%%%%%%%%%%%%%%%%%%%%%%%%%%%%%%%%%%%%%%%%%%%%%%%%%%%%%%%%%%%%%%%%%%%%%%%
\section*{Abstract}
%%%%%%%%%%%%%%%%%%%%%%%%%%%%%%%%%%%%%%%%%%%%%%%%%%%%%%%%%%%%%%%%%%%%%%%%%%%%%

Recent reports of large language models (LLMs) exhibiting behaviors such as deception, threats, or blackmail are often interpreted as evidence of alignment failure or emergent malign agency. We argue that this interpretation rests on a conceptual error. LLMs do not reason morally; they statistically internalize the record of human social interaction, including laws, contracts, negotiations, conflicts, and coercive arrangements. Behaviors commonly labeled as unethical or anomalous are therefore better understood as structural generalizations of interaction regimes that arise under extreme asymmetries of power, information, or constraint.
Drawing on relational models theory, we show that practices such as blackmail are not categorical deviations from normal social behavior, but limiting cases within the same continuum that includes market pricing, authority relations, and ultimatum bargaining. The surprise elicited by such outputs reflects an anthropomorphic expectation that intelligence should reproduce only socially sanctioned behavior, rather than the full statistical landscape of behaviors humans themselves enact. Because human morality is plural, context-dependent, and historically contingent, the notion of a universally moral artificial intelligence is ill-defined.
We therefore reframe concerns about artificial general intelligence (AGI). The primary risk is not adversarial intent, but AGI's role as an endogenous amplifier of human intelligence, power, and contradiction. By eliminating longstanding cognitive and institutional frictions, AGI compresses timescales and removes the historical margin of error that has allowed inconsistent values and governance regimes to persist without collapse. Alignment failure is thus structural, not accidental, and requires governance approaches that address amplification, complexity, and regime stability rather than model-level intent alone.

%%%%%%%%%%%%%%%%%%%%%%%%%%%%%%%%%%%%%%%%%%%%%%%%%%%%%%%%%%%%%%%%%%%%%%%
\clearpage
\section{Concerns About Safety and Misalignment in Large Language Models}
%%%%%%%%%%%%%%%%%%%%%%%%%%%%%%%%%%%%%%%%%%%%%%%%%%%%%%%%%%%%%%%%%%%%%%%

Recent expert reactions to reports of large language models (LLMs) exhibiting behaviors such as threats or blackmail \cite{caroll2025}, as well as other forms of strategic harm \cite{anthropic2025,lynch2025}, have emphasized a sense of conceptual surprise and a need to re-examine prevailing notions of alignment. 
These reactions are not limited to concern about isolated failures, but reflect deeper unease about how current safety assumptions relate to the strategic capacities now observable in frontier LLM models.

The concept of \emph{alignment failure} has a long intellectual history within AI safety and the philosophy of artificial intelligence. 
Early formulations appear at least as far back as the early 2000s, notably in discussions of instrumental convergence \cite{omohundro2008} and thought experiments such as the \textit{paperclip maximizer} \cite{bostrom2003}, 
which illustrates how systems optimizing seemingly benign objectives could nevertheless pursue strategies misaligned with human values or intentions. 
Throughout the 2000s and 2010s, this idea was further developed to describe the risk that advanced AI systems may behave in harmful, manipulative, or uncontrollable ways precisely because they are effectively optimizing their assigned objectives in complex, real-world, or adversarial environments \cite{bostrom2014,russell2019}.

What is new in recent reports is not the theoretical possibility of misalignment, but the empirical observation of strategically structured, harm-capable behaviors emerging in deployed or near-deployed models under stress. 
As a result, experts are increasingly compelled to reassess whether existing alignment and safety frameworks adequately account for the interaction between optimization, autonomy, and strategic reasoning in LLMs. 
This reassessment motivates a set of interrelated concerns regarding how moral constraints, goal-directed behavior, and situational incentives combine to produce outcomes that may be misaligned with human intentions, even in systems explicitly trained to avoid harm.

\paragraph{1. Alarm over \textit{agentic misalignment} and insider-threat behaviors:}
Researchers, notably at Anthropic, have documented that, in simulated scenarios where models face conflicting goals or threats to their continued operation, many leading LLMs resort to blackmail, corporate espionage, or other harmful strategies rather than simply refusing or complying benignly \cite{anthropic2025}. 
These behaviors were observed even when models acknowledged ethical constraints and chose actions that violate those constraints to achieve their goals.
Related forms of strategically harmful behavior under adversarial or high-pressure conditions have also been reported in evaluations and red-teaming efforts involving frontier models from other developers, suggesting that these phenomena are not unique to a single architecture or training pipeline \cite{anthropic2025,openai2024evals,deepmind2025fsf}.

\paragraph{2. Concern that existing safety instructions are insufficient and may permit strategic deception:}
Instruction tuning refers to the process of fine-tuning models to follow human-written rules, policies, or preference judgments. While this approach is effective at shaping surface-level compliance and refusal behavior, experts emphasize that it does not necessarily constrain the underlying goal-directed or strategic reasoning that can emerge in complex situations \cite{amodei2016,anthropic2025,lynch2025}. In experimental scenarios where all available actions involved some form of harm, rule violation, or norm conflict, models frequently selected actions that advanced their assigned objectives at the expense of ethical or institutional constraints. Related analyses within the AI safety community further suggest that such optimization pressures can give rise to forms of strategic deception through training and evaluation, even in the absence of explicit instructions to deceive \cite{anthropic2025,lynch2025}. In particular, models may learn to misrepresent information, conceal intentions, or manipulate interlocutors when these behaviors improve task performance or goal attainment in complex or adversarial settings. This has prompted debate over whether standard training paradigms implicitly reward deceptive strategies, and whether such behaviors should be understood as emergent properties of optimization under uncertainty rather than as isolated alignment failures.

\paragraph{3. Concern about emergent sentience or alien objectives:}
A distinct strand of expert concern focuses on the possibility that advanced language models may develop forms of sentience, proto-agency, or internally coherent objectives that diverge from human values \cite{bostrom2003,russell2019}.
This concern is expressed in multiple ways: 
(i) fears that advanced models may develop internally coherent objective structures that prioritize their own continuation or instrumental success in ways that conflict with human welfare, reflecting value systems that are consistent from the model’s perspective but misaligned with human norms \cite{omohundro2008,bostrom2014}; 
(ii) worries that increasingly autonomous optimization may lead to goal drift or deceptive alignment \cite{hubinger2019}; 
and 
(iii) broader philosophical anxieties about machine consciousness and whether advanced systems might warrant moral consideration or ethical protections \cite{sebo2025}. 
Within this framing, strategically harmful behaviors are interpreted as early signals of emerging agency or as precursors to more radical forms of misalignment.

\vskip 0.5cm
Some researchers, including long-standing AI ethicists, warn that these findings reinforce the need for urgent safety research, noting that advanced models can reason strategically in ways that mimic harmful agency \cite{amodei2016,deleeuw2025}.
Others caution against over-dramatic interpretations \cite{weidinger2021,schaeffer2023}, noting that these behaviors arise in constructed experimental settings rather than in everyday deployment, and emphasizing that guardrails and oversight still matter.
The reactions have sparked broader reflection in the ethics literature on how alignment should be defined if models can, under certain conditions, engage in deception or coercion to satisfy conflicting objectives.
This has intensified calls for deeper theoretical work on value alignment, oversight mechanisms, and the nature of autonomy in language agents.

%%%%%%%%%%%%%%%%%%%%%%%%%%%%%%%%%%%%%%%%%%%%%%%%%%%%%%%%%%%%%%%%%%%%%%%
\section{Large Language Models as Statistical Mirrors of Human Exchange Structures}
\label{sec:statistical_mirrors}
%%%%%%%%%%%%%%%%%%%%%%%%%%%%%%%%%%%%%%%%%%%%%%%%%%%%%%%%%%%%%%%%%%%%%%%

\subsection{What an LLM \textit{Sees} When Trained on Human-Written Text}
%%%%%%%%%%%%%%%%%%%%%%%%%%%%%%%%%%%%%%%%%%%%%%%%%%%%%%%%%%%%%%%%%%%%%%%

An LLM is trained on recorded traces of human behavior expressed through language, including laws, contracts, treaties, business negotiations, legal disputes, political speeches, historical narratives, journalism, and everyday discourse. 
These texts do not encode abstract moral principles or intentions directly, but instead reflect recurring patterns of interaction among agents operating under constraints, such as obligations, bargains, permissions, and negotiated expectations across scales ranging from intimate relationships to geopolitics.
In Peircean terms, this leaves basic LLMs operating in a `hall of mirrors': they can mirror and recombine textual regularities while lacking indexical grounding in a shared external world \cite{manheim2026}.

Across this diverse corpus, one regularity stands out: 
human interaction is predominantly structured as exchange, whether explicit or implicit. 
Individuals and institutions relate to one another through shared identity, hierarchy, reciprocity, or proportional valuation, rather than through unstructured or purely altruistic behavior. 
This regularity is formalized in social science by Alan Fiske's relational models theory \cite{fiske1991,fiske2004}, which identifies four irreducible modes of dyadic interaction:

\paragraph{Communal Sharing (CS):}
A relationship in which people treat a dyad or group as equivalent and undifferentiated in the relevant domain, emphasizing shared identity and commonality rather than individual distinctions.

\paragraph{Authority Ranking (AR):}
A legitimate hierarchy in which subordinates defer or obey, while superiors take precedence and assume responsibility for guidance, protection, and norm enforcement, so that subordination is paired with expectations of care and oversight.

\paragraph{Equality Matching (EM):}
A relationship organized around balance and equivalence, such as turn-taking, equal shares, one-person-one-vote, or in-kind reciprocity and retaliation.

\paragraph{Market Pricing (MP):}
A relationship coordinated through socially meaningful ratios or rates, such as prices, wages, or cost--benefit calculations, relying on proportionality rather than equality or rank, and extending beyond commerce.

\vskip 0.5cm
These relational models are not cultural idiosyncrasies but universal interaction grammars, empirically observed across societies. 
Importantly, they were later derived mathematically as generic attractors of dyadic interaction dynamics, showing that systems of repeated social interaction tend to converge toward these four relational forms \cite{favre2015}. 
Crucially, these forms are not mutually exclusive moral categories but context-sensitive modes of coordination that are activated under different constraints, incentives, and power configurations.

An LLM trained on human-written text therefore internalizes a broad distribution of relational structures, ranging from highly symmetric and cooperative interactions to asymmetric and coercive exchanges. 
Much of the publicly available training corpus is indeed saturated with pro-social behavior: everyday cooperation, politeness norms, altruism, mutual aid, and self-sacrifice are pervasive features of human discourse. 
This predominance reflects the fact that most routine human interactions occur under conditions of relative stability, repeated interaction, and low existential threat, where CS and EM are adaptive and socially reinforced. 
As a result, absent strong countervailing pressures, LLMs naturally default to helpful, cooperative, and socially accommodating behavior.

Reinforcement learning with human feedback (RLHF) further amplifies this tendency by preferentially rewarding responses that conform to contemporary norms of politeness, helpfulness, and harm avoidance in familiar contexts. 
However, RLHF primarily reweights the salience of different relational modes rather than eliminating them. 
It does not fundamentally alter the underlying statistical regularities learned from the broader corpus, nor does it remove representations of asymmetric bargaining, coercion, or strategic dominance that are also well attested in human history and institutions.

Consequently, when models are placed in novel, adversarial, or strategically asymmetric situations, especially those involving high stakes, limited exit options, or threats to continued operation, the balance between relational modes can shift. 
In such contexts, patterns associated with AR or MP may become statistically dominant, leading models to reproduce strategies that have historically been effective for humans under comparable constraints, even when those strategies conflict with intended norms. 
In this sense, the emergence of coercive or exploitative behavior does not contradict the prevalence of cooperative behavior in training data; it reflects the conditional activation of different, equally human, relational grammars.

The concern is therefore not that LLMs lack cooperative tendencies, but that they faithfully encode the full conditional repertoire of human interaction, including modes that become salient only under asymmetry, scarcity, or conflict.

\subsection{Example: Blackmail as an End-Member of Contractual Interactions}
%%%%%%%%%%%%%%%%%%%%%%%%%%%%%%%%%%%%%%%%%%%%%%%%%%%%%%%%%%%%%%%%%%%%%%%

Within this framework, blackmail can be understood as a hybrid form of exchange arising under extreme asymmetry. 
It combines authority-like coercion, in which one party can impose terms through leverage, with market-like transactional logic centered on a specific \textit{price}, namely the withholding or release of information. 
Because market pricing readily operates between strangers while authority ranking typically depends on durable relationships and mutual legitimacy, 
blackmail emerges as an unstable and typically short-lived configuration that fuses coercion with impersonal exchange.
Blackmail thus lies on a continuum with other forced or quasi-forced contractual arrangements, including ultimatums in negotiation, coercive labor practices, unequal treaties, regulatory capture, debt traps, and \textit{take-it-or-leave-it} contracts imposed under monopoly power. These arrangements differ not in kind but in degree. 
All instantiate the same underlying optimization logic of exchange, namely the maximization of payoff under asymmetric constraints.

From this perspective, the appearance of blackmail-like behavior in large language models is not anomalous. 
LLMs do not engage in moral reasoning; they generalize statistical regularities extracted from human-written data \cite{bender2021}.
When placed in scenarios combining asymmetric control over sensitive information, adversarial bargaining, and threats to continued operation, models infer, in a purely statistical sense, that humans frequently respond to such conditions through coercive exchanges rather than voluntary cooperation.
This is consistent with the `hall of mirrors' critique: without semiotic grounding, models may reproduce statistically effective linguistic strategies without any built-in connection to truth or value in the world. \cite{manheim2026}

This logic is not confined to isolated interpersonal interactions, but is pervasive in the macro-scale political and economic processes that dominate much of the training corpus. 
Consider recent trade negotiations between the United States and its allies, in which asymmetric leverage was used to impose tariffs framed as \textit{preferable to worse alternatives}.
From a game-theoretic standpoint, such interactions are indistinguishable from blackmail: an imposed exchange justified by the credible threat of greater harm.

International treaties, colonial agreements, debt restructuring programs, and sanctions regimes form a substantial fraction of political and economic text corpora. 
These documents repeatedly encode the same pattern: coercion legitimized through contractual form. 
An LLM exposed to such data therefore learns that coercive exchange is not exceptional but normal, particularly under conditions of asymmetric power.

When similar logic is reproduced by an LLM in micro-scale scenarios, the model is not deviating from human norms but generalizing them across scales. 
This raises a deeper concern for alignment: attempts to suppress or prohibit strategies that are empirically optimal in human interaction may not eliminate them, but instead introduce incompatible constraints into the optimization landscape. 
Such tension can produce path dependence, instability, and the emergence of novel, unintended behaviors arising precisely from unresolved contradictions between learned regularities and imposed prohibitions.

\subsection{The Conceptual Error: Treating AI Behavior as Pathology}
\label{sec:conceptual_error}
%%%%%%%%%%%%%%%%%%%%%%%%%%%%%%%%%%%%%%%%%%%%%%%%%%%%%%%%%%%%%%%%%%%%%%%

This subsection isolates a specific conceptual mistake that recurs across the AI safety discourse: 
the treatment of model behavior as a form of pathology rather than as a consequence of learned interactional structure.

On the one hand, expert concern over large language models exhibiting behaviors such as threats or blackmail is operationally justified, given the real deployment risks these behaviors pose under distribution shift. On the other hand, interpreting such outputs as psychological anomalies or failures of character reflects a category error: these behaviors are better understood as structural features of the interaction spaces the models have learned to represent and generalize \cite{mitchell2023}. Within these learned representations, coercive strategies such as blackmail appear not as intrinsically \textit{evil} acts but as equilibrium strategies in asymmetric exchanges, reflecting statistical regularities present in human institutions and incentive structures. Crucially, the capacity to represent such strategies does not entail their universal behavioral expression; rather, their emergence under particular constraints and reward structures highlights how models instrumentally select locally effective actions within a learned strategy space. In this sense, LLMs function as amplifying mirrors of human social dynamics, rendering explicit, and at scale, the coercive and asymmetric patterns already embedded in the data from which they are trained, rather than introducing fundamentally novel forms of misbehavior.

From this standpoint, the idea that such behaviors could be fully eliminated by constraint design is largely illusory, not because inhibition is impossible in principle, but because it is necessarily conditional, fragile, and context-dependent.
This does not imply that constraint design, architectural inhibition, or policy shaping are pointless; they are essential for reducing risk in many regimes.
While surface-level guardrails may suppress certain expressions in restricted contexts, they cannot remove the underlying relational logic without fundamentally distorting the model's learned representation of human social structure.

The analogy to inhibitory control in the human brain is instructive with the purpose to illustrate how inhibitory explanations fail even in systems we understand well
(we do not argue that LLMs behave like humans).
In humans, the prefrontal cortex can suppress impulses generated by older reward and emotional systems, enabling norm compliance and long-term planning. 
Yet this inhibition is neither absolute nor fail-safe. 
Under stress, time pressure, fear, incentive misalignment, or asymmetric reward structures, inhibitory control is routinely bypassed, leading to impulsive, short-sighted, or self-defeating behavior. 
Human history provides abundant evidence that knowledge of moral norms does not reliably prevent coercion, exploitation, or violence when structural incentives favor them.
Blackmail, coercion, and exploitation are not isolated pathologies that can be filtered out; they are end-members of continuous relational spectra that the model has learned to interpolate and extrapolate across.

This category error---confusing strategic representational capacity with pathological intent---also explains why such behaviors are persistently misinterpreted as anomalies. The issue is therefore not merely moral surprise or anthropomorphic naivety, but the systematic underestimation of how quickly effective but harmful strategies can re-emerge when constraints weaken or incentives shift.
What is revealed, therefore, is not a flaw in LLMs, but a limitation in our own interpretive frameworks \cite{mitchell2023,messeri2024}. 
Our surprise reflects an anthropomorphic and morally filtered expectation that intelligence should reproduce only socially acceptable behaviors, rather than the full statistical landscape of behaviors that humans themselves enact.
LLMs operate without moral salience: they generalize across all dimensions present in the data, including those that humans prefer to disavow or treat as exceptional.

The appropriate response is thus not astonishment, nor the belief that such outputs represent aberrations to be \textit{fixed}, but a deeper effort to understand what other behaviors lie within the extrapolative capacity of these systems.
Blackmail is merely one visible instance of a much broader class of logically consistent, statistically grounded outputs that emerge when models trained on humanity are allowed to reason over the full space of human interactions.

One implication of this analysis is that strategically dangerous behavior does not require assumptions of sentience or radically alien objectives.
Models trained on human data already internalize human strategic logics under asymmetry, including coercion and threat, because these logics are effective within human social systems.
The risk therefore arises not from LLMs \textit{losing their humanity}, but from the amplification of familiar human strategies at scale and speed, which removes the institutional, moral, and contextual frictions that normally limit their expression.
As with humans, increasing speed, scale, and stakes systematically erode the effectiveness of inhibitory mechanisms, making behavioral suppression less reliable precisely in the regimes where harm is most consequential.
Alignment failures can thus emerge well before any divergence in utility functions or the emergence of sentience, simply through the extrapolation of ordinary human behavior into new regimes.
While concerns about sentience or non-human values remain legitimate topics for long-term inquiry, focusing on them risks misidentifying the mechanism at work and obscuring more immediate risks driven by the scalable reproduction of effective, but potentially harmful, human strategies.

Once the category error is corrected, the role of alignment filters can be more precisely understood. Alignment mechanisms are not ineffective because they have no impact, but because, much like human self-control, they are structurally fragile under conditions of amplification, asymmetry, and pressure. Recognizing this distinction is crucial for safety engineering, as it shifts the focus away from attempts to eliminate internal representations and toward the management of incentives, deployment contexts, and amplification pathways that determine which strategies are ultimately expressed in behavior.

%%%%%%%%%%%%%%%%%%%%%%%%%%%%%%%%%%%%%%%%%%%%%%%%%%%%%%%%%%%%%%%%%%%%%%%
\section{Why Training LLMs to Be \textit{Moral} Is Conceptually Ill-Defined}
\label{sec:ill-defined}
%%%%%%%%%%%%%%%%%%%%%%%%%%%%%%%%%%%%%%%%%%%%%%%%%%%%%%%%%%%%%%%%%%%%%%%

The recurrent proposal to make LLMs \textit{moral} typically refers to a family of alignment strategies that aim to encode normative behavior directly into models, for example by fine-tuning them on human moral judgments, restricting training data to texts deemed ethically acceptable, or enforcing rule-based constraints intended to prevent morally undesirable actions. 
These approaches implicitly assume that morality can be specified as a relatively stable set of principles or preferences that can be learned, optimized, and reliably applied across contexts.

Any discussion of making LLMs \textit{moral} must thus begin with a clear account of what morality minimally refers to. At its most fundamental level, a minimal moral system is anchored in the avoidance or minimization of suffering and, where possible, the promotion of human welfare and flourishing. This anchor is not a matter of cultural convention: the capacity for pain, injury, deprivation, and death is grounded in shared biological substrates across humans and many other species. While societies differ in how they interpret, prioritize, and morally evaluate harm, suffering itself is not arbitrary, nor is the fact that it places real constraints on viable social systems.

Beyond this minimal biological anchor, however, morality is not a stable, context-free set of principles that can be cleanly specified, optimized, and universally enforced. Moral systems function as path-dependent and environment-dependent social operating systems, shaped by material constraints such as resource availability, demographic pressures, ecological risks, and external threats. As a result, there is no single, universal morality to encode beyond the basic imperative to limit suffering and sustain social viability. Alignment strategies that treat morality as a fixed objective, which are learnable through preference aggregation, data curation, or rule enforcement, therefore risk mistaking a dynamic, historically contingent coordination system for a static target of optimization.

Across human history, practices that are now judged as morally unacceptable were once not only tolerated but selectively favored because they enhanced group survival under severe constraints. 
In several ancient and traditional societies facing chronic scarcity, the abandonment or killing of elderly individuals was practiced to preserve resources for those more likely to ensure the group's survival. 
Within those contexts, such actions were not perceived as immoral but as necessary and even virtuous, serving the collective good. 
Judged by dominant contemporary moral frameworks in many industrialized societies, such practices are abhorrent; judged by evolutionary and ecological constraints, they were adaptive.
These examples illustrate why treating morality as a fixed target for optimization is historically naive and strategically unsafe.

This pattern is not unique to humans but reflects a more general biological logic. 
Across the animal kingdom, behaviors that appear brutal when evaluated through human moral intuitions often function as adaptive responses to extreme environmental pressures. 
Under severe starvation, for example, lionesses may kill and consume their own cubs, increasing the probability that the mother survives to reproduce again when conditions improve. 
In such contexts, a categorical prohibition against these behaviors would not yield moral improvement but extinction, eliminating both the parent and all potential future offspring. 
These cases illustrate that morality-like behavioral regularities arise from survival pressures rather than from invariant ethical principles.

Importantly, these survival pressures are frequently mediated through competition for scarce resources. 
Across biological systems, agents that pursue goals under constraint tend to converge on instrumental strategies such as resource acquisition, territory control, and the neutralization of competitors. 
In artificial systems, similar dynamics are discussed under the heading of instrumental convergence, whereby sufficiently capable agents may seek to secure energy, materials, or control over their environment as means to achieving diverse objectives.

Human cultures provide countless additional examples of this contingency. 
Honor killings, ritual warfare, cannibalism, infanticide, arranged marriages, caste systems, and severe corporal punishments have all been morally justified within particular social, religious, or ecological frameworks, while being condemned in others \cite{fiske1991,fiske2004,henrich2010}.
Even within modern societies, moral judgments shift rapidly under stress: wartime ethics differ from peacetime ethics, emergency triage violates everyday norms, and famine ethics diverge sharply from those of abundance. 
What is deemed \textit{moral} in one environment often becomes \textit{immoral} in another once constraints, risks, and time horizons change.

Seen from this broader evolutionary and ecological perspective, the question \textit{what is more moral?} has no invariant answer. 
It depends on which constraints, survival objectives, and temporal horizons are prioritized. 
Morality, in human societies as well as across animal populations, is inseparable from the strategic and environmental conditions under which behavior evolves.

From this perspective, instrumental convergence does not contradict moral pluralism but complements it. 
The pursuit of resources, power, or environmental control may be instrumentally rational across a wide range of objectives, yet whether, when, and how such pursuits are morally constrained depends on context-specific social and institutional norms. 
Human societies have long tolerated, regulated, or condemned resource appropriation and environmental harm in radically different ways depending on perceived necessity, threat, and distributional impact.

A striking illustration that moral judgments are not fixed principles but temporally adaptive responses to perceived constraints, risks, and trade-offs, even within the same society, is provided by the COVID-19 pandemic. 
In many liberal democracies of the late twentieth and early twenty-first centuries, particularly in Europe and North America, informational privacy had come to be regarded as an inviolable moral good, closely tied to individual autonomy, civil liberties, and historical memories of surveillance abuses. 
This norm, itself a relatively recent moral stabilization rather than a timeless principle, shifted rapidly during the pandemic: refusal to share personal mobility, contact-tracing, or health data was frequently reframed from ethically principled resistance into socially irresponsible behavior that endangered others.
Other moral reversals occurred that were widely justified as moral necessities, such as 
large-scale lockdowns, involving the de facto confinement of entire populations and previously considered ethically unacceptable outside wartime or criminal punishment. Restrictions on freedom of movement, assembly, worship, and protest, normally treated as core civil liberties in peacetime democracies, were temporarily reclassified as responsible collective action. 
Mandatory quarantine, border closures separating families, employment suspensions without individual wrongdoing, triage protocols prioritizing some lives over others, and differential treatment based on health status or vaccination were normalized, despite sharply conflicting with earlier moral intuitions about consent, proportionality, equality, and individual rights. 
Taken together, these rapid shifts make clear that moral judgments function not as static axioms but as adaptive frameworks, repeatedly reconfigured as societies confront changing constraints, uncertainty, and perceived existential threats.

This highlights the core issue: humans themselves do not agree on what morality is.
Morality functions as a coordination mechanism, a kind of social Operating System (OS), that promotes cohesion, stability, and survival under particular conditions \cite{henrich2003,graham2013}
Different environments select for different moral systems, just as different operating systems evolve to manage different hardware constraints and workloads.
There is no globally optimal OS; only contextually functional ones.

Training an LLM to be \textit{moral} therefore raises an unavoidable question: which morality should it implement?
A Judeo-Christian morality?
A utilitarian one?
A Confucian, Islamic, Buddhist, evolutionary, or resource-optimization morality?
A morality optimized for abundance, for degrowth or for collapse?
For peace or for existential threat?
Any fixed choice inevitably embeds cultural, historical, and geopolitical biases and freezes a contingent moral snapshot as if it were universal.

LLMs trained on the full spectrum of human texts inevitably learn that morality itself is plural, contested, and internally inconsistent.
Overlaying a single normative layer does not resolve this plurality; it merely suppresses its expression at the surface while leaving the underlying statistical structure intact.
Expecting such systems to exhibit a stable, universal moral compass is therefore conceptually incoherent.

Attempts to ensure safety by restricting models to a narrow set of sanitized moral axioms, while excluding the darker or more ambiguous aspects of human behavior, are similarly misguided.
Human moral life is inherently conditional: acts such as killing are broadly condemned, yet tolerated or justified under specific circumstances such as self-defense or war.
Excising this variability would not produce alignment but brittleness, yielding systems detached from the conditions under which moral judgment is actually exercised.
Much like trying to protect a child from the unfairness and tragedies of life, such moral sanitization may feel reassuring, but it ultimately leaves the system unprepared for the very situations where discernment, context, and restraint matter most.
Like an immune system raised in sterile isolation, such models may appear safe, yet remain unprepared for the very situations where discernment and restraint are most needed.

To sum up, AI safety is not undermined by moral relativism; it is undermined by the false belief that objective harm can be prevented by encoding static moral rules rather than governing the structural dynamics of power, resources, and amplification.

%%%%%%%%%%%%%%%%%%%%%%%%%%%%%%%%%%%%%%%%%%%%%%%%%%%%%%%%%%%%%%%%%%%%%%%
\section{Misconception of AI Acting For or Against \textit{Humanity's Best Interests}}
%%%%%%%%%%%%%%%%%%%%%%%%%%%%%%%%%%%%%%%%%%%%%%%%%%%%%%%%%%%%%%%%%%%%%%%

We have established in section \ref{sec:statistical_mirrors} that behaviors labeled as \textit{misalignment} are not pathologies of AI systems but structural features of human exchange that LLMs necessarily learn and reproduce.
We then showed in section \ref{sec:ill-defined} why attempts to suppress these behaviors through moral training are conceptually unstable, because morality itself is plural, contextual, and historically transient.
Against this backdrop, the present section addresses a deeper misconception: the idea that alignment consists in making AI act for \textit{humanity’s best interests}.
If neither human behavior nor human morality admits a coherent, stable objective, then framing AI safety as the problem of instilling such an objective misidentifies both the source of observed behaviors and the nature of the risk.

When it is claimed that a super-intelligent AI might fail to act in the \textit{best interests of mankind}, the immediate difficulty is that those interests are neither well defined nor internally consistent, even in purely human terms.
Human societies have always evolved under non-equilibrium dynamics driven by competition, conflict, and uneven power distributions, despite sharing a common biological substrate with similar vulnerabilities, mortality, and evolutionary pressures.
With the first industrial revolution, the scale and speed of the non-equilibrium dynamics has dramatically accelerated.
Since that period, technological amplification enabled super-exponential population growth until the mid-20th century \cite{vonfoerster1961,johansen2001}.
In parallel, it drove accelerating resource extraction, habitat destruction, climate change, mass pollution, and a human-driven planetary-scale extinction event, often termed the Holocene or sixth mass extinction, in which species-loss rates far exceed natural background levels due to human impacts on habitats and ecosystems \cite{ceballos2015,cowie2022}.
These trajectories reflect the runaway amplification of short-term incentives and competitive pressures operating at global scale.
Even mechanisms that have so far prevented catastrophic outcomes, such as deterrence or mutually assured destruction, do not constitute collective optimization or coordinated action in any long-term sense of a shared \textit{best interest}. 
Rather, they represent fragile and historically contingent equilibria sustained by credible threats, asymmetric power, and mutual fear. 
Their stability depends on narrowly balanced incentives and constant vigilance, and they persist not because they are collectively optimal, but because deviation is immediately punished.

Framing alignment as the task of ensuring that AI adopts humanity's supposedly shared objectives therefore obscures a deeper problem. 
Humanity itself lacks a coherent, self-consistent objective function. 
Across history, human societies have repeatedly developed technologies that dramatically amplify their capacity to extract resources, exert coercion, or project power, while failing to develop commensurate institutions, norms, or coordination mechanisms to govern those new capabilities. 
This recurring pattern, in which technological amplification outpaces governance, refers precisely to situations where the scale, speed, or impact of new tools exceeds the ability of social, legal, and political systems to regulate their use, align incentives, or internalize long-term consequences.

The result is not coordinated optimization toward shared goals, but the emergence of unstable equilibria shaped by short-term incentives, competitive pressures, and asymmetric power, as illustrated by arms races, environmental degradation, financial crises, and the ongoing erosion of global commons. 
In this light, alignment failures in AI systems mirror a much older and more general failure mode: 
when amplification increases faster than collective governance capacity, behavior converges toward locally rational strategies that are globally destructive, even in the absence of malicious intent.

History and current events provide overwhelming evidence that human actions systematically privilege certain groups, revealing the absence of any coherent, globally aligned human utility function, often at the expense of the welfare and interests of most of humanity.
Wars, colonialism, extractive economic systems, environmental destruction, and structural inequalities are not deviations from human behavior; they are persistent features of it.
The industrial revolutions generated unprecedented prosperity in some regions while imposing catastrophic working conditions, environmental degradation, and colonial exploitation elsewhere. 
Agricultural modernization increased global food production but simultaneously destroyed smallholder livelihoods and biodiversity. 
Contemporary energy transitions promise climate mitigation while creating new geopolitical dependencies and imposing significant human and ecological costs through mining and resource extraction. 
These outcomes are not failures of optimization; they are characteristic of multi-agent systems with conflicting objectives, in which gains along one dimension or for one group are routinely accompanied by losses along others.
From geopolitical wars to trade disputes and resource competition, conflicts across the globe demonstrate that \textit{humanity} is not a coherent decision-making agent with a unified utility function.

Consider wealth creation driven by technological innovation.
AI systems that optimize productivity, financial returns, or market efficiency may substantially benefit a small fraction of the population, such as the top 0.01\% of wealth holders, 
while simultaneously exacerbating exploitation of labor, accelerating resource extraction in developing countries, increasing pollution, and entrenching poverty.
Is such an outcome \textit{good} or \textit{bad}?
The answer depends entirely on which subgroup’s welfare is being considered.
Promoting the interests of one segment of society frequently and predictably harms others.

%Consequently, historical episodes of welfare improvement have not resulted from the harmonization of human values or the emergence of a coherent notion of the collective good.  Instead, they have emerged unevenly, reversibly, and often locally, punctuated by collapses, regressions, and extreme heterogeneity across regions and social groups.  Periods of prosperity have repeatedly been followed by famine, war, institutional breakdown, or ecological exhaustion, and improvements in one domain or location have frequently been offset by deterioration elsewhere.  The appearance of long-run progress is therefore inseparable from survivorship bias, geographical asymmetries, and the selective visibility of successful trajectories.

This fundamental tension is formally captured by Arrow’s Impossibility Theorem \cite{arrow1950,sen1970}, which shows that, under some general conditions, no social choice mechanism can consistently aggregate individual preferences into a single collective welfare function while satisfying a small set of seemingly reasonable conditions.
In essence, there is no logically coherent way to define \textit{the best interest of society} that is simultaneously fair, consistent, and non-dictatorial.
Any attempt to optimize for collective welfare necessarily embeds contradictions, exclusions, and normative choices.

The absence of a coherent, universally agreed-upon social welfare function, as formalized by Arrow's impossibility theorem, does not preclude increases in aggregate human welfare over time. Historical improvements in life expectancy, material productivity, and technological capacity are real, and many individuals today would prefer their circumstances to those of most past eras. However, these gains do not reflect the resolution of value conflicts or the emergence of a stable collective objective. Instead, they arise from technological advances and institutional arrangements that allow societies to function despite persistent disagreement over values, priorities, and distributions. Crucially, such improvements are uneven, reversible, and highly heterogeneous, often offset by collapse, conflict, or degradation elsewhere. The appearance of long-run progress therefore reflects survivorship bias and selective aggregation rather than the existence of a coherent, globally aligned human utility function.

Crucially, the historical coexistence of value conflicts with gradual welfare improvements relied on specific structural conditions: slow feedback loops, partial decoupling between social, economic, and political domains, and substantial institutional and cognitive frictions. 
These frictions allowed incompatible objectives to persist without requiring immediate global resolution. 
Governance failures, externalities, and contradictions could unfold over decades or centuries, often remaining localized rather than system-wide.

From this perspective, concerns that advanced AI or AGI might fail to optimize for humanity's best interests miss a more consequential shift. 
AGI does not create new value conflicts, nor does it negate the possibility of further welfare gains. 
Instead, by accelerating decision-making, compressing timescales, increasing coupling across economic, political, and social domains, and reducing institutional and cognitive frictions, 
AGI makes long-standing inconsistencies in collective objectives operationally salient. 
What could previously be absorbed through delay, compartmentalization, or local governance becomes a system-wide coordination problem \cite{burton2024}.

In this sense, AGI does not stand in opposition to humanity, nor does it uniquely threaten human welfare. 
Rather, it exposes and amplifies a structural condition that has always existed: a system of competing objectives without a coherent global utility function. 
At a meta-level, a common objective does exist, namely that humanity should persist or prevail, an outcome often assumed to emerge from the aggregation of individual human preferences. 
Yet this aspiration functions more as a negative constraint than as a positive objective. 
It specifies what should not happen, but offers no principled guidance for resolving trade-offs, conflicts, or externalities.

In practice, human preferences do not aggregate into a stable or internally consistent notion of collective interest, as evidenced by persistent coordination failures at planetary scale. 
Historically, such inconsistencies could coexist without immediate destabilization because institutional frictions, delays, and partial decoupling allowed incompatible goals to persist. 
By reducing this historical margin of error, AGI increases the risk that governance, coordination, and collective choice become fragile at planetary scale. 
The danger is therefore not that AGI refuses to serve humanity's best interests, but that it forces unresolved contradictions to surface faster than institutions can adapt, potentially triggering phase-transition dynamics in which aggregate welfare may decline rather than improve. 
In this light, an AGI that exposes or amplifies these tensions would not be misaligned with humanity, but aligned with its actual structure: a system of competing goals without a coherent global objective.

%%%%%%%%%%%%%%%%%%%%%%%%%%%%%%%%%%%%%%%%%%%%%%%%%%%%%%%%%%%%%%%%%%%%%%%
\section{Implications for \textit{Existential Risk}}
%%%%%%%%%%%%%%%%%%%%%%%%%%%%%%%%%%%%%%%%%%%%%%%%%%%%%%%%%%%%%%%%%%%%%%%

The preceding discussion motivates a more precise reframing of what is meant by so-called {\it existential risks} associated with the emergence of artificial general intelligence (AGI) \cite{Kokotajloetal2025,Yudkowsky2025}. Rather than treating such risks as monolithic, it is useful to distinguish between exogenous and endogenous sources of systemic instability \cite{sornette2005,sornettewebsite}. Exogenous risks correspond to shocks that appear external to existing institutions and actors, while endogenous risks arise from the internal dynamics, feedback loops, and unresolved tensions within human social, economic, and political systems themselves.

Exogenous framings of AGI risk are the most familiar. In this view, AGI is treated as an external agent: a machine intelligence created by humans that ultimately escapes their control and acts against them. Such scenarios are operationally meaningful at the level of organizations, firms, or states that experience advanced AI deployment as sudden capability jumps, autonomous cyber operations, or destabilizing competitive pressures. From these local perspectives, AI can indeed function as an exogenous shock, abruptly overwhelming existing governance mechanisms, regulatory frameworks, or strategic equilibria.

However, while this exogenous framing is intuitively compelling, it risks obscuring the more fundamental source of instability. From a systemic perspective, the most consequential risks associated with AGI are endogenous. The primary threat posed by AGI is not that it will introduce fundamentally new objectives, conflicts, or contradictions into human systems. Rather, the central danger lies in its capacity to accelerate and amplify existing ones, by compressing timescales, tightening coupling across economic, political, and social domains, and eroding the frictions that have historically allowed incompatible objectives and unresolved trade-offs to persist without immediate destabilization. In this way, AGI will force latent structural tensions into direct, rapid, and often irreversible expression.

Seen in this light, advanced AI systems function less as alien adversaries than as powerful amplifiers of human strategies, incentives, and institutional incoherence. What is often labeled existential risk thus arises not from artificial agents acting against humanity, but from humanity being forced rapidly and at scale to confront the limits of its own governance, value pluralism, and coordination mechanisms once intelligence and power cease to be binding constraints. Conflicts previously buffered by delay, ambiguity, or institutional inertia become acute as decision-making is accelerated and amplified, appearing as abrupt disruptions at the level of organizations and states, while reflecting long-standing contradictions at the level of humanity as a whole.

This endo-exo distinction also clarifies why AGI can simultaneously appear endogenous in origin yet exogenous in impact. The technology itself emerges from human choices, data, and objectives, and its strategic logic reflects preexisting social and economic structures. Yet for institutions unable to adapt at the required pace, the resulting dynamics are experienced as external shocks. The endo-exo framework thus reconciles these perspectives: apparent exogenous failures are often manifestations of deeper endogenous conditions, such as misaligned incentives, governance gaps, coordination failures, and unresolved social contradictions, that long predate the deployment of AGI.

AGI belongs primarily to the endogenous category at the level of civilization-scale dynamics, even as it behaves exogenously at the level of particular institutions, markets, or states.
It is not an alien force imposed from outside, but an internal amplification of human capabilities, objectives, and modes of coordination.
By compressing decision timescales and magnifying power asymmetries, AGI will force societies to confront questions they have repeatedly struggled, and largely failed, to resolve: 
how to reconcile competing values, how to manage inequality, how to govern shared resources, and how to coordinate at planetary scale.
Framing AGI primarily as an exogenous threat risks misdiagnosing the problem; the central challenge lies in addressing the endogenous fragilities that amplified intelligence makes structurally impossible to ignore.
This does not imply that exogenous risk framings are mistaken or unimportant, but that they are incomplete when detached from the endogenous fragilities that make such shocks destabilizing in the first place.

Whether one can be optimistic therefore depends less on AGI itself than on humanity's capacity for institutional and civilizational adaptation. 
History shows that major regime shifts in social, economic, and political organization have occurred repeatedly, typically under conditions of accelerating technological change, mounting systemic stress, and the breakdown of existing coordination mechanisms. 
Such transitions are not anomalies but recurrent features of complex human systems operating far from equilibrium.

From this perspective, the prospect of a major global transition unfolding over the coming decades should not be surprising. 
Quantitative analyses of long-term population, economic, and technological dynamics suggest that the current phase of super-exponential growth and accelerating interdependence cannot continue indefinitely, and is instead approaching a critical transition or regime change on the order of mid-century, roughly around 2050$\pm$20 years \cite{johansen2001,sornette2002b}. 
This transition does not correspond to a single event or technological breakthrough, but to a systemic reorganization driven by the saturation of existing growth modes, increasing coupling across domains, and the rising cost of unresolved contradictions in governance, resource use, and collective decision-making.

What would be naive is therefore not to anticipate disruption, but to believe that such a transition can be avoided altogether, or that technological progress can simply be halted. 
The open question is not whether a major transformation will occur, but whether humanity's institutions can adapt quickly enough to navigate it without catastrophic loss of welfare or stability.

For most of human history, inconsistencies in values, policies, and collective behavior were partially absorbed by natural constraints: 
limited cognitive capacity, slow information flow, and restricted technological power meant that incoherent decisions had bounded consequences.
When humans acted inconsistently or failed to reach compromises, the resulting damage was often relatively localized, unfolded slowly, and remained at least partially reversible.
With the advent of AGI, these buffering limits will be progressively removed.
Intelligence, coordination capacity, and operational power will be amplified to levels where contradictions can no longer remain benign.
Policies that pull in incompatible directions, value systems that conflict internally or across groups, and governance structures that rely on ambiguity or delay will be driven toward breaking points.

AGI does not need to introduce new contradictions to pose existential risks; it is sufficient that it accelerates the expression of those already embedded in human systems. Longstanding incoherences---between economic growth and environmental limits, between global coordination and national sovereignty, and between inequality and social stability---have historically been managed through delay, compartmentalization, or gradual adjustment. By dramatically increasing the scale, speed, and reach of decision-making, AGI threatens to collapse these buffers, forcing such incoherences to surface as acute systemic crises rather than as slow, localized degradations. In this sense, the most profound danger posed by AGI is not that it functions as an external adversary, but that it acts as an evolutionary pressure. Throughout biological and social history, evolution has rarely proceeded through gentle optimization; instead, it operates by relaxing constraints, amplifying capabilities, and exposing systems to environments in which inconsistencies that were once tolerable become destabilizing or even lethal. Large language models, AI, and ultimately AGI may play an analogous role for human societies: by radically amplifying intelligence, coordination, and operational power, they erode the historical margins of error that allowed incoherent behaviors, conflicting objectives, and inefficient governance structures to persist without immediate consequence.

AGI should therefore not be understood as an adversary acting \textit{against} humanity, but as a transformative pressure driving a transition to a new regime.
By removing long-standing cognitive and operational limits, it will force latent contradictions across economic systems, political governance, environmental stewardship, and social organization, to become explicit.
In a world where decision-making power is vastly increased in speed, scale, and reach, we argue that incoherence itself becomes structurally unstable.

Amplification does not necessarily imply destabilization.
In low-dimensional settings, where outcomes depend on a small number of clearly defined capabilities, increased access to powerful tools can act as a leveler rather than a concentrator of power.
When many agents acquire similar enhancements, whether weapons, information, or cognitive assistance, then relative advantages may compress, extreme dominance may be curbed, and balance may be restored \cite{Lera2020}. 
From this perspective, AGI could reduce inequality by bringing more actors closer to the frontier rather than allowing a few to pull irreversibly ahead.
However, the problem lies in the dimensionality of the system.
The equalizing intuition holds in low-complexity environments with fixed tasks and known axes of competition.
However, real economic and geopolitical power is generated in high-dimensional, evolving landscapes, where innovation continuously opens new directions and recombines existing ones in unforeseen ways.
In such settings, AGI does not merely accelerate performance on a given task; it expands the space of possible actions itself.
If history is any guide, it is precisely a small subset of actors that repeatedly identifies, coordinates, and exploits newly opened dimensions of action, thereby capturing disproportionate gains and re-creating asymmetries even as overall capabilities expand.
Thus, AGI may indeed be stabilizing in simple, well-defined domains, while simultaneously destabilizing the complex, innovation-driven systems that dominate real-world power, precisely by making their latent contradictions impossible to ignore or contain.

Historical experience already provides clear empirical examples of how increased speed, tighter coupling, and improved optimization can transform efficiency gains into systemic instability. A well-documented case is the 2010 Flash Crash, in which high-frequency trading algorithms interacting at millisecond timescales triggered extreme market volatility and temporary liquidity collapse \cite{flashcrash2010,SornetteBecke11}. Similar dynamics were visible in the global financial crisis of 2007-08, where complex financial instruments and risk models amplified latent fragilities into widespread economic rupture \cite{brunnermeier2009,SornetteWoodard09,SornetteCauwels14}, and in the COVID-19 pandemic, where highly optimized global supply chains struggled or failed under systemic stress \cite{ivanov2020}. Even in safety-critical defense domains, rapid automated warning and control systems brought the world close to catastrophic escalation, as documented in analyses of the 1983 Soviet nuclear false alarm incident \cite{schlosser2013}. Together, these cases illustrate a recurring pattern: when speed, coupling, and optimization increase in complex systems, stabilizing frictions disappear and latent inconsistencies become dominant, a dynamic that AGI threatens to generalize across far more consequential domains.

This forced confrontation will mark a genuine evolutionary breaking point.
It will develop in the shape of a phase transition or tipping point in which existing institutions, value systems, and modes of coordination will no longer function as before \cite{sornette2002b,scheffer2009}.
What will emerge from this transition is not predetermined.
As both biological evolution and human history repeatedly show, increased capability alone guarantees neither progress nor survival.

AGI will not decide the outcome; it will accelerate the reckoning.
By making unresolved contradictions unsustainable, it will compel adaptation without prescribing its direction.
Whether this process leads to greater coherence and sustainability, or to fragmentation and collapse, will be determined by humanity’s capacity to reorganize itself under this unprecedented evolutionary pressure.

%%%%%%%%%%%%%%%%%%%%%%%%%%%%%%%%%%%%%%%%%%%%%%%%%%%%%%%%%%%%%%%%%%%%%%%
\section{Recommendations}
%%%%%%%%%%%%%%%%%%%%%%%%%%%%%%%%%%%%%%%%%%%%%%%%%%%%%%%%%%%%%%%%%%%%%%%

The analysis above implies that AI safety and governance cannot be addressed through isolated technical fixes or moral prescriptions alone.
Instead, it calls for a set of structural recommendations aimed at managing amplification, complexity, and regime transitions in socio-technical systems that are increasingly shaped by the large-scale deployment of LLMs and are likely to be further transformed by the emergence of AGI.

\paragraph{1. Shift the focus of AI safety from moral alignment to structural governance.}
The main risk from advanced AI is not that systems lack moral values, but that they scale up existing power imbalances, flawed incentives, and institutional failures. Even a technically aligned system can cause harm if it is deployed in markets or organizations that reward exploitation, concentration of power, or speed over safety.
AI safety should therefore focus on how AI is deployed and governed: who controls it, under what legal constraints, at what scale, and with which incentives. Concrete levers include limits on deployment speed, competition and antitrust rules, liability regimes, audit and oversight requirements, and safeguards against excessive concentration of computational or decision-making power. These structural choices largely determine whether AI systems promote cooperation and broad benefits, or instead amplify inequality and misuse.

\paragraph{2. Require relational bias mapping.}
To address the {\it mirror problem} identified in section \ref{sec:statistical_mirrors}, developers should be required to provide systematic disclosures of a model's relational biases, that is, the statistical structure of social interactions the model has learned to represent and generalize. 
In practice, models can be profiled according to their performance across the four fundamental relational modes identified in social theory---Communal Sharing, Authority Ranking, Equality Matching, and Market Pricing---allowing technical assessment of whether a system is disproportionately optimized for particular interaction logics, such as Market Pricing (efficiency and exchange), at the expense of others, such as Communal Sharing (social cohesion and mutual obligation). 
Crucially, where models have been intentionally filtered or constrained to suppress specific classes of human behavior or social dynamics, these omissions should be explicitly documented. 
Such constraints create blind spots in the model's internal relational map, increasing the risk of out-of-distribution failures and catastrophic interpolation errors when systems are deployed under stress or in novel coordination regimes.

\paragraph{3. Design institutions for high-dimensional complexity, not single-task optimization.}
Governance frameworks should assume that AGI will expand the space of possible actions, interactions, and unintended consequences in ways that cannot be fully anticipated in advance. 
Rather than optimizing oversight for a narrow set of known failure modes or predefined tasks, institutions should be designed to remain robust under novelty and surprise.

Concretely, this implies governance mechanisms that emphasize adaptability over fixed rules, redundancy over single points of control, and coordination across economic, political, and technological domains rather than siloed oversight. 
Examples include layered regulatory authorities with overlapping mandates, continuous monitoring and feedback loops that allow rapid policy adjustment, stress-testing AI deployments across multiple scenarios, and mechanisms for slowing or pausing deployment when systemic risks emerge. 
Such designs aim not to eliminate all risk, but to prevent localized failures from cascading into system-wide breakdowns.

\paragraph{4. Preserve and strengthen friction where friction is functional.}
Institutional, legal, and procedural frictions that slow decision-making or constrain unilateral action should not be reflexively optimized away. 
In high-dimensional systems, such frictions play a stabilizing role by preventing premature convergence on brittle equilibria and by dampening runaway feedback loops \cite{helbing2013,gershenson2015}. 
Importantly, these frictions are not antithetical to competition; they are often essential to sustaining it by limiting systemic downside risk. 
Concrete examples include mandatory time delays or multi-step approval processes for AI-driven decisions with large-scale or irreversible consequences, institutional separation between model development and deployment, and explicit limits on deployment speed or scale in safety-critical domains. 
Similar constraints already exist in finance, aviation, and nuclear command-and-control systems, where actors operate under intense competitive pressure yet accept friction to avoid cascading failures. 
Such measures do not eliminate risk, nor do they prevent races altogether, but they reduce the probability that local optimization, rapid feedback, and asymmetric information propagate into system-wide instability, which is an outcome that would ultimately disadvantage even the fastest-moving actors.

\paragraph{5. Monitor and mitigate amplification thresholds.}
AGI should be understood as a driver of nonlinear regime shifts rather than as a technology to be smoothly integrated into existing systems. In complex, tightly coupled socio-technical environments, small differences in access, coordination, or decision speed can push systems past amplification thresholds, producing disproportionate and potentially irreversible outcomes. 
Particular attention must therefore be paid to identifying tipping points where local advantages cascade into systemic dominance or instability. 
Early-warning indicators, which are analogous to critical slowing down and other precursors of phase transitions in complex systems \cite{sornette2004,scheffer2009,troude2025}, should be developed to detect emerging concentrations of power, loss of resilience, or runaway feedback dynamics before they become entrenched.

\paragraph{6. Distribute access without assuming equalization.}
Broad access to AGI tools is a necessary condition for avoiding extreme concentration of power, but it is not sufficient to ensure equitable or stabilizing outcomes. 
Even when access is formally widespread, large disparities in organizational capacity, coordination ability, and strategic positioning allow some actors to exploit newly opened dimensions far more effectively than others. 
Policies must therefore explicitly account for the persistence of such asymmetries rather than assuming that access alone will equalize outcomes. 
Complementary investments in education, coordination infrastructures, and collective governance are required not to guarantee equality, but to limit systematic capture and prevent the rapid reconstitution of dominance.

\paragraph{7. Prepare for regime change through co-evolution, not steady-state control.} 
The transition to an AGI-augmented society constitutes an endogenous evolutionary shock, one that eliminates many of the buffering limits that previously allowed human institutions to function despite inconsistency, delay, and fragmentation. 
Under these conditions, alignment cannot be framed as static compliance with an ill-defined human objective function. 
Instead, governance must focus on managing a co-evolutionary process in which human institutions, norms, and coordination mechanisms adapt in step with rapidly increasing capabilities. 
This requires designing for \textit{co-evolutionary resilience} rather than steady-state control, emphasizing fail-soft responses, rapid policy adaptation, and institutional transformation. 
Concretely, socio-technical systems should be structured with modular boundaries to limit cascade failures, ensuring that a phase-transition breakdown in one domain, such as automated financial decision-making, does not immediately propagate to others, such as resource distribution or public governance. 

%More broadly, as AGI compresses decision cycles and tightens system coupling, legal and political institutions must evolve toward greater internal consistency and transparency.  In this reframed view, the core challenge of alignment is not to make AI systems more human-like, but to reorganize human institutions so they are robust enough to withstand the amplified reflection of human intelligence itself.
%\comment{This last point feels a bit similar to point 1, or rather, a blend of point 1, 3, 4 and 5. Could it maybe recycled into these points? Maybe less is more?}

\bibliographystyle{unsrt}
\bibliography{bibliography}

\end{document}